\documentclass[usletterpaper, 10pt, ieeeconf]{ieeeconf}      
\usepackage{geometry}
\newgeometry{left=72pt,right=54pt,top=54pt,bottom=54pt}
\usepackage[bookmarks=false]{hyperref}
\usepackage[noadjust]{cite}
\usepackage{graphicx}
\usepackage{subcaption}
\usepackage{float}
\usepackage{amsfonts,amsmath,amssymb,array} 

\IEEEoverridecommandlockouts                              

\title{\LARGE \bf 3D Reconstruction \& Assessment Framework \\based on affordable 2D Lidar}

\author{Xueyang Kang$^{1}$ Shengjiong Yin$^{2}$ Yinglong Fen$^{1}$
\thanks{*This work was not supported by any organization}
\thanks{$^{1}$Xueyang Kang and Yinlong Fen are Master student of Electrical and Information Engineering, Technical University of Munich, Munich D-80333, Germany{\tt\small alexander.kang@tum.de}}
\thanks{$^{2}$Shengjiong Yin is with the Smart Monitoring Laboratory, Tongji University, Shanghai 201804, China}
}

\begin{document}

\begin{titlepage}
\vspace*{1cm}

This paper has been accepted for publication in 2018 International Conference on \textit{Advanced Intelligent Mechatronics (AIM)}, Auckland, New Zealand.

\textcircled{c} 2018 IEEE. Personal use of this material is permitted. Permission from IEEE must be obtained for all other uses, in any current or future media, including reprinting/republishing this material for advertising or promotional purposes, creating new collective works, for resale or redistribution to servers or lists, or reuse of any copyrighted component of this work in other works.
\end{titlepage}

\maketitle
\thispagestyle{empty}
\pagestyle{empty}

\begin{abstract}
Lidar is extensively used in the industry and mass-market. Due to its measurement accuracy and insensitivity to illumination compared to cameras. It is applied onto a broad range of applications, like geodetic engineering, self-driving cars or virtual reality. But the 3D Lidar with multi-beam is very expensive, and the massive measurements data cannot be fully leveraged on some constrained platforms. The purpose of this paper is to explore the possibility of using cheap 2D Lidar off-the-shelf, to perform complex 3D reconstruction, moreover, the generated 3D map quality is evaluated by our proposed metrics at the end. The 3D map is constructed in two ways, one way in which the scan is performed at known positions with an external rotary axis not parallel to the intrinsic rotary axis of Lidar. The other way, in which the 2D Lidar for mapping and another 2D Lidar for localization are placed on a trolley, the trolley is pushed on the ground arbitrarily. The generated maps by different approaches are converted to octomaps uniformly before the evaluation. The similarity and difference between two maps will be evaluated by the proposed metrics thoroughly. The whole mapping system is composed of several modular components. A 3D bracket was made for assembling of the Lidar with a long range, the driver and the motor together. A cover platform made for the IMU and 2D Lidar with a shorter range but high accuracy. The software is stacked up in different ROS packages.
\end{abstract}

\section{INTRODUCTION \& RELATED WORK}
Lidar has been one of the most anticipated sensors in recent years. It emits light pulse with unique identity, and correlates the light bouncing off the surface with the original signal. The depth information to the object can be measured in three ways, the time of flight(ToF) measurement, phase-shift measurement, and triangulation. At present, the most affordable 2D Lidar sensors in the market are with a single laser beam. The upper part of the device containing sender and receiver is mounted onto the motor shaft, to get a 360-degree view.

Building 3D scan based on 2D laser requires additional dimension, such as, (a) adding a fixed rotary axis to extend 2D device. (b) 2D scanner mounted onto the mobile platform, generating 3D data along moving. Each of them has its own pros and cons. Another issue is how to evaluate the quality of 3D map. Since it is difficult to obtain the ground truth in reality, the metrics are only devised to rate the relative quality among compared maps. To evaluation process, implementation directly on 3D point cloud is very computationally expensive, hence the reasonable solution should utilize existing efficient storage structure, like multi-dimensional tree search.

Many existing methods to construct 3D map via 2D scanner is to extend dimension, like a tilting angle can be introduced into scanning process for a hand-held device ~\cite{3D scanning}, some open source projects about 3D reconstruction via Lidar~\cite{3D kit} have already borrowed from this approach. Obviously, this method adds to the complexity of the hardware, furthermore, the coordination of two rotary axes is complicated.

The trivial work involved in the fusion of different sensor measurements, requires to calibrate the different outputs, and synchronize the different data streams~\cite{synchronization}. For positioning, the traditional method depends on IMU to implement odometry, but the outputs of accelerometer need to be integrated twice, consequently the drift error will accumulate over the time, the estimation from wheel encoder also suffers from this problem. The alternative is to use camera to implement visual odometry(VO) ~\cite{Visual odom1},~\cite{Visual odom2}. The accuracy of the current mature VO algorithms is superior to IMU. However, if the lighting condition varies too much abruptly, this method will not work. But Lidar based Simultaneous Localization and Mapping(SLAM) can deal with these problems and achieve robust performance.

The main 2D SLAM algorithms in ROS community includes: "gmapping"~\cite{gmapping}, an improved Rao-Blackwellized algorithm based on the particle filter, depending both on the Lidar and the odometer, in "gmapping", the Lidar outputs for measurement model and the odometer outputs for motion model executed iteratively in succession; "HectorSLAM"~\cite{hector} based on 2D grid map, the algorithm uses scan match to find the optimal transform and estimate the new position. The objective for optimization is a function of the occupancy probability. Each grid cell in the 2D map is registered along with the occupancy probability. The continuous model is approximated through the bi-linear interpolation of the probability, Hector is only dependent on the Lidar, it can register multiple maps with different resolutions and retrieve them on demand; the newly released algorithm "cartographer"~\cite{cartographer} with a several meters' drift error on a kilometer's trajectory, the loop closure detection and pose optimization have also been added into "cartographer" to further imrpove map consistency, and the pruned search is introduced to speed up the match search. 
\newgeometry{left=54pt,right=54pt,top=54pt,bottom=55pt}
Some graph-based SLAM algorithms utilize the existing topological structure in the world to optimize the map~\cite{topological_slam}. The mathematical model behind SLAM algorithm is recursive probability update, as stated in~\cite{probabilistic robotic},~\cite{bayes_network}.

Traditionally, the root-mean-square error~\cite{RMSE} can be employed in evaluation when the ground truth is available. In computer vision, the IoU metric~\cite{IoU} calculates the common pixels in two images to make a pixel-wise comparison. But 3D point array produced by Lidar,is an inefficient memory management way to be used for evaluation, the octomap~\cite{octomap} is a tree based structure, in which the endpoints are represented by the cubes, a type of 3D voxels~\cite{3d_voxel}.

\section{SYSTEM OVERVIEW}
Two types of sensors are adopted, IMU and two 2D Lidar sensors, Sweep Scanse and Rplidar. The whole system is composed of three parts: the localization part integrates Rplidar and 9 axes IMU; the 3D scanner part, includes Sweep Scanse, stepper motor and motor driver; the laptop as back-end on which the fusion algorithm, the post-processing pipeline, and visualization process run. All sensors were installed onto the 3D printed kits to be protected.

\subsection{Hardware}
\begin{figure}[thpb]
     \centering
      \framebox{\parbox{3.0in}{\includegraphics[scale = 0.32]{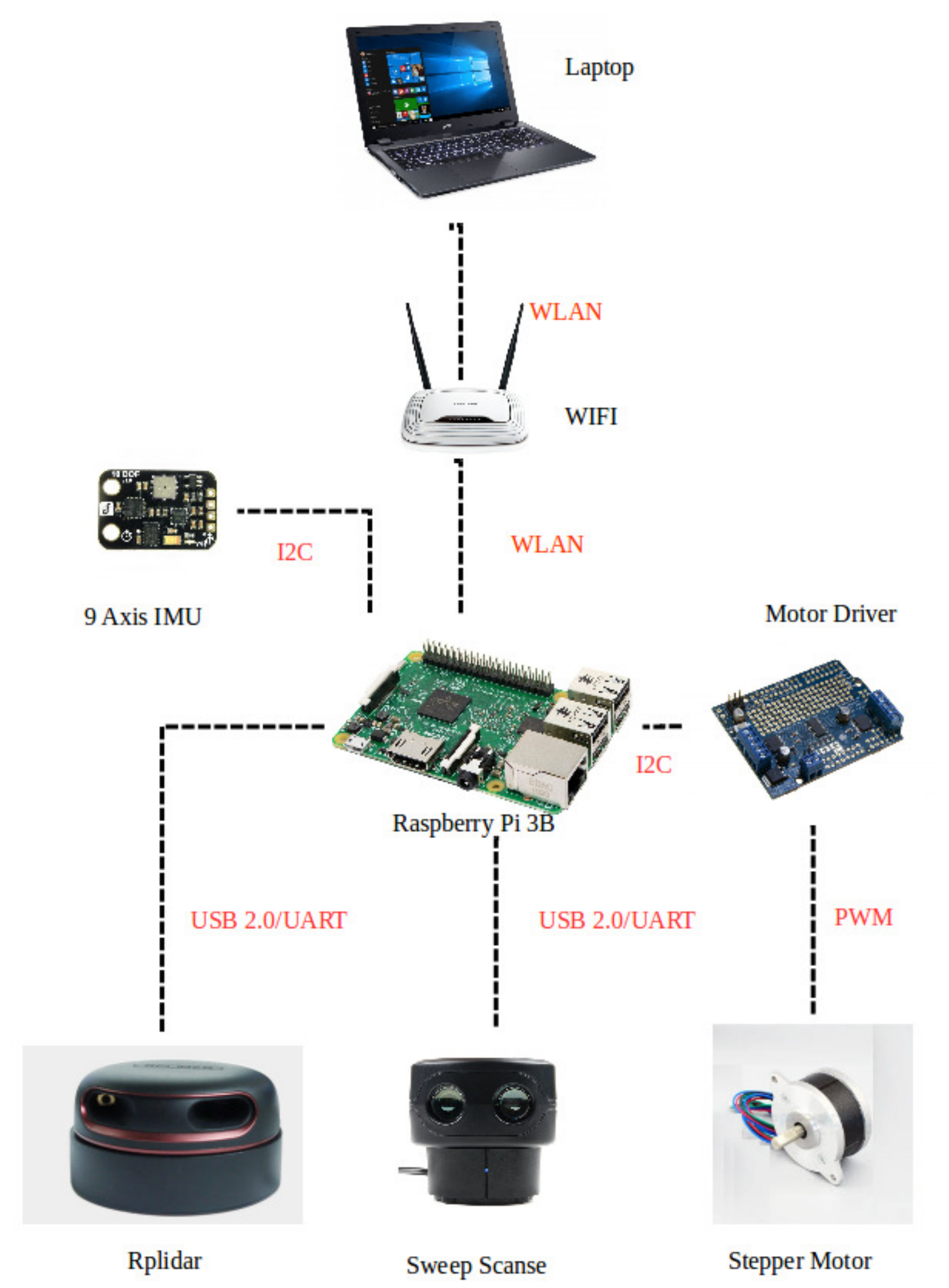}}}    
      \caption{System overview}
      \label{fig:outline}
   \end{figure}

The system components and transmission protocols are presented in Fig\ref{fig:outline}. The Raspberry Pi 3B is responsible for collecting the measurements from sensors as front-end, while the laptop serves as back-end. In fusion mode, the measurements from Sweep Scanse and Rplidar, as well as the measurements from gyroscope, accelerometer, magnetometer, are sampled by Raspberry Pi 3B, and then wirelessly transmitted to laptop. The 3D scan at stationary locations only requires the measurements from Sweep Scanse. and the stepper motor will provide the additional dimension. Particularly, the stepper motor is directly driven by the dedicated PWM signals from driver board. The communication between Raspberry Pi and laptop is through WLAN. The measurements from IMU and the controlling commands for the stepper motor are transmitted via I2C, but their transmission directions are different. The two Lidar sensors are connected to Raspberry Pi by USB cables, without extra power required.

The parameters provided by the Lidar suppliers are listed in the following table. Rplidar is with short range, but higher resolution, while Sweep Scanse has opposite features. This is mainly due to their different measurement principles, Rplidar uses triangulation measurement, while Sweep Scanse uses ToF. Each 2D Lidar costs about 300 to 500 dollars.

\begin{table}[thpb]
\caption{Specifications comparison}
\label{table_sepcs}
\begin{center}
\begin{tabular}{|c|c|c|c|}
\hline
  &\textbf{Sampling rate(samples/s)}&  
  \textbf{Range(m)} &\textbf{Frequency(Hz)}\\
  \hline
  \textbf{Rplidar}& 4000 & 0.15-6 & 1-11 \\
  \hline
  \textbf{Sweep}& 1000 & 0.1-40 & 1-10\\
   \hline
\end{tabular}
\end{center}
\end{table}

\begin{figure}[thpb]
     \centering
      \includegraphics[scale = 0.20]{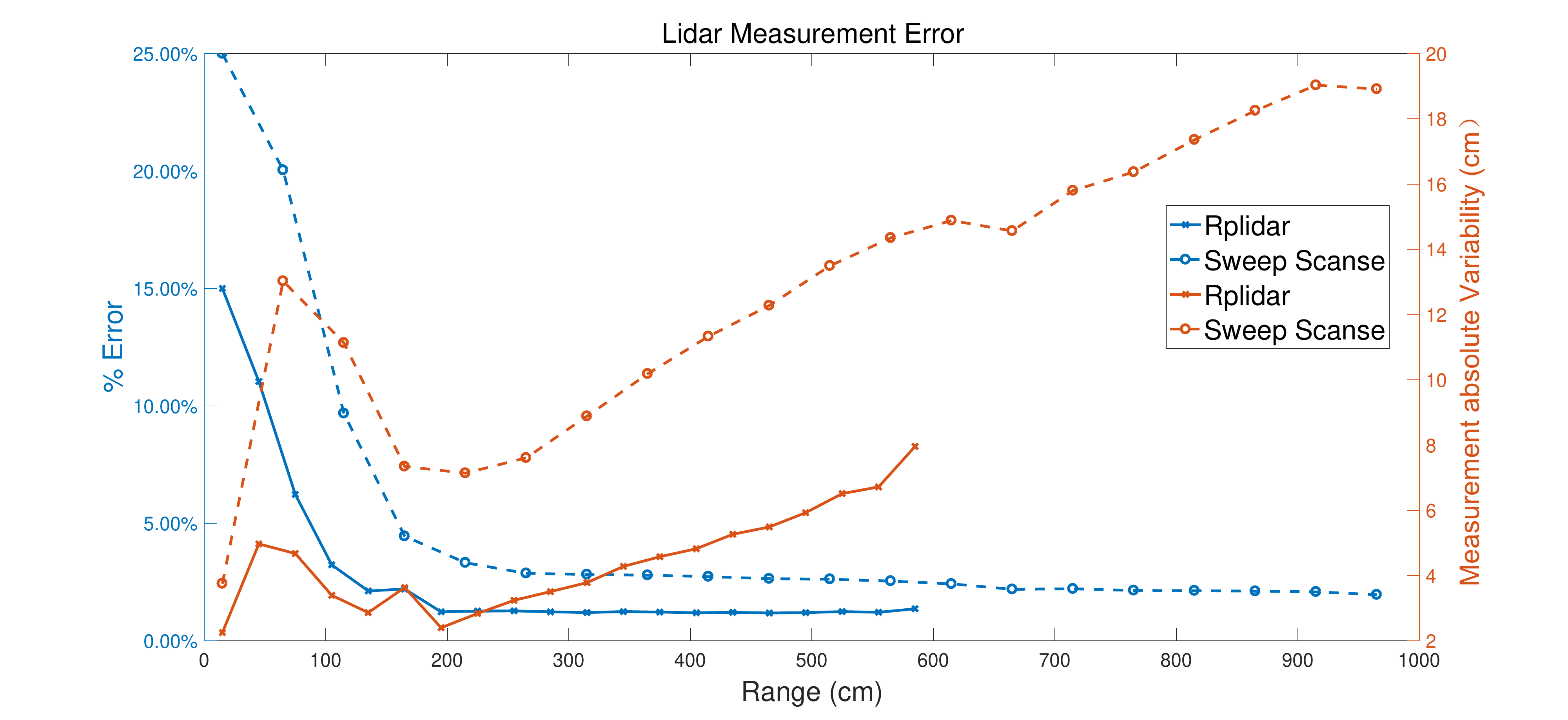}    
      \caption{Absolute and relative errors}
      \label{fig:error}
   \end{figure}

Fig. \ref{fig:error} shows the measurement errors from experimental tests. For both Lidars, the relative errors in the left y-axis drop drastically at the distance ranging from 1.5 to 2.0 meters, finally level off at about 2\%, as the measuring distance increases. But the absolute error in right y-axis initially fluctuates, then increases gradually. In general, both the relative and absolute error of Sweep Scanse are greater than those of Rplidar. It is worth noting that, the maximal range 40 meters claimed by Sweep Scanse inventor is not real. The possibility of getting valid measurements is very small at distance above 10 meters according to test, consequently the 10 meters is adopted as the range at software level for valid measurement.

\subsection{Software}
\begin{figure}[thpb]
     \centering
      \includegraphics[scale = 0.24]{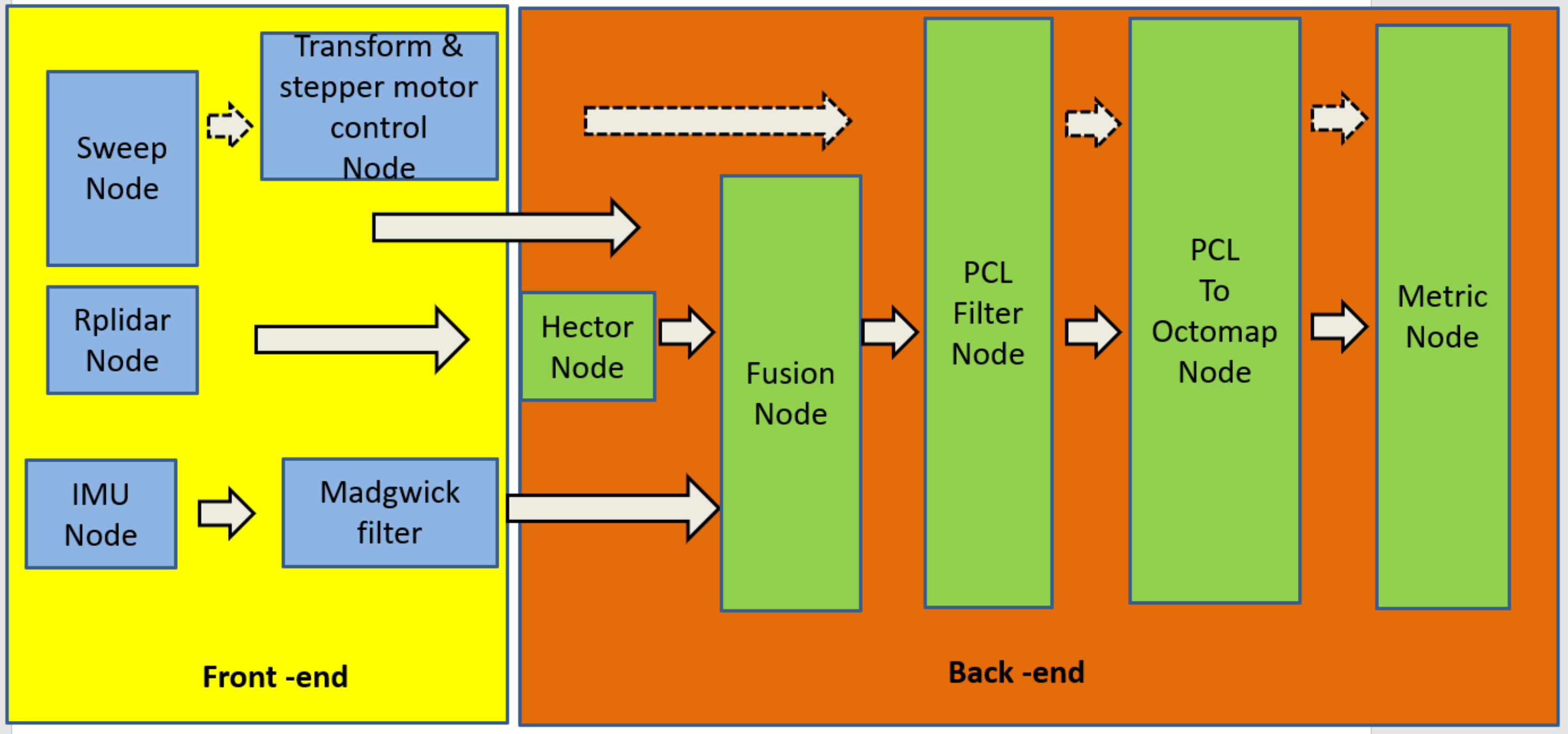}    
      \caption{Node work-flow}
      \label{fig:work-flow}
   \end{figure}

The whole software is stacked on Ros framework hosted on \textit{Ubuntu} system. The fusion algorithm transforms measurements from local frame into a same global frame, and matches the different data-streams based on their sampling time. Then the transformed point cloud is passed to the post-processing pipeline to reject outliers, and converted to octomap. Lastly, the evaluation process runs offline to compare the two maps constructed by different methods or under different hardware settings. Fig. \ref{fig:work-flow} contains all ROS nodes in the system, the arrow denotes the topic passed to the subscriber node. Left block contains the driver nodes of sensors, which work on Raspberry Pi, right block contains the nodes running on laptop. Especially, the measurements from IMU are processed by "Madgwick filter"~\cite{madgwick} on Raspberry Pi, because the IMU outputs are sampled at high rate, the filter implemented at front-end can avoid the transmission latency. The ROS nodes connected by the dashed arrows, are regarding the 3D reconstruction at static locations. The remaining nodes pertaining to the 3D reconstruction along movement. The "Fusion Node" fuses all the measurements from IMU, Rplidar, and Sweep Scanse together to generate 3D map. The pipeline from "PCL Filter Node" to "PCL to Octomap Node", all the way up to "Metric Node", is to post-process the point cloud and convert it into the octomap. 

\section{3D RECONSTRUCTION}
3D reconstruction methods based on 2D Lidar are mainly divided into two types, as mentioned previously: 3D scan at static positions, or incremental 2D scan along movement.

\subsection{3D Reconstruction At Known Locations}
The figure below shows the entire 3D scanner device after assembling the motor, the driver, and Raspberry Pi into the 3D printing suite.

\begin{figure}[bhtp]
\centering
\includegraphics[scale=.08]{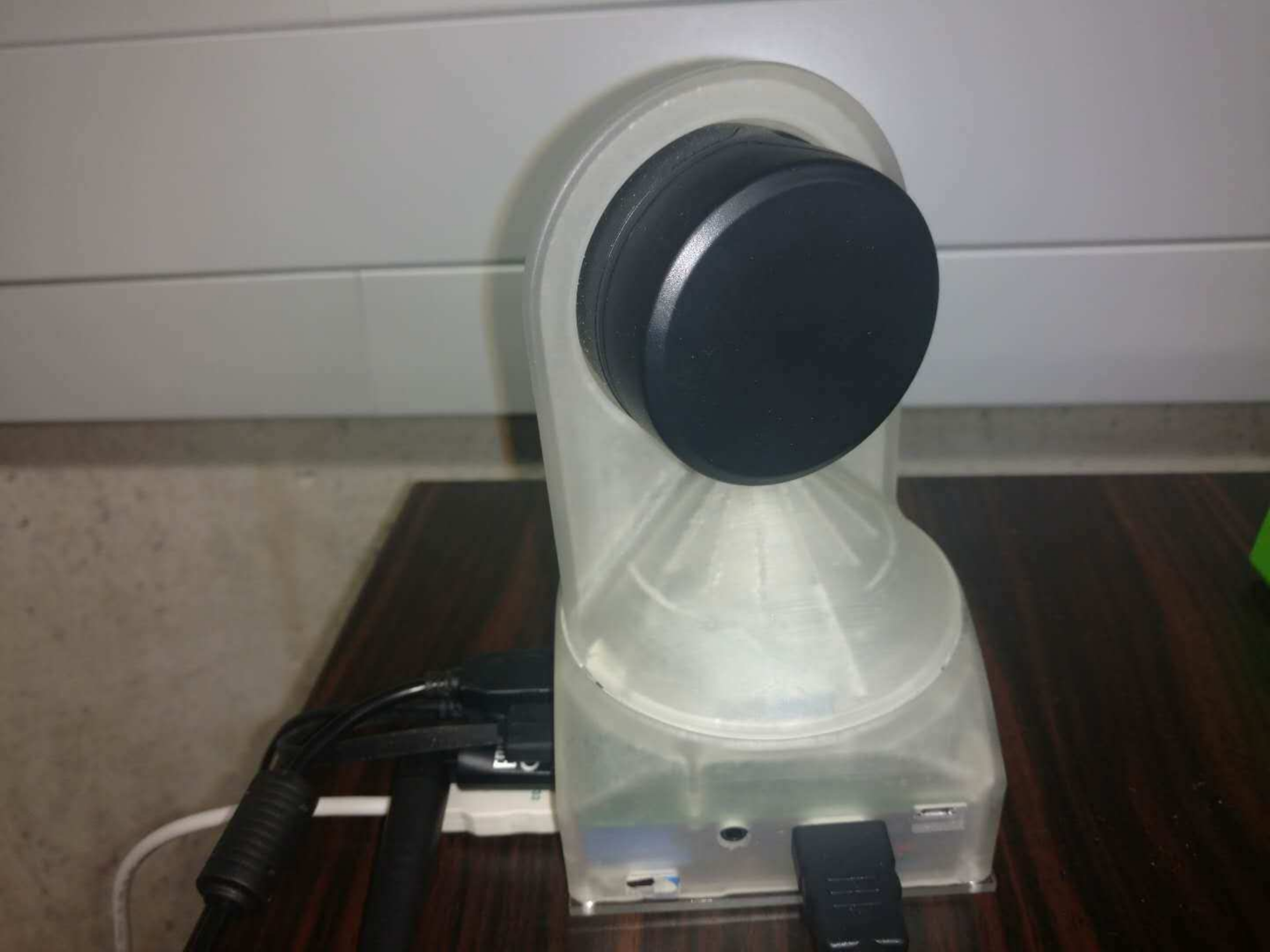}
\caption{3D scan device}
\label{fig:scan-device}
\end{figure}

The static position $({x}_p, {y}_p)$ on the ground is provided by optical instrumentation with high precision, the orientation $\phi$ of the device is determined by the stepper motor rotation angle. Each of the measurement pair, range $L$ and bearing $\theta$ in Sweep Scanse's local frame can be transformed to a 3D point $({x}_{i}, {y}_{i}, {z}_{i})$ in global frame. But the bearing $\theta$, ranging from $0^\circ$ to $180^\circ$, is transformed by the current yaw angle $\phi$, for the bearing, ranging from $180^\circ$ to $360^\circ$, the accumulative yaw orientation $\phi$ of the device is updated after stepper motor rotation by $\delta\phi$, this is because when the beam is pointing to the bottom of the device, measuring is blocked, hence this interval can be taken advantage of, to rotate the upper part of the device to the new orientation.

\begin{footnotesize}
\begin{equation}\label{eq:trans}
\left[\begin{matrix} 
    {x}_{i}\\
    {y}_{i}\\
    {z}_{i}\\
\end{matrix} \right] 
 = \begin{cases}
 \left[\begin{matrix} 
    {x}_p - Lsin(\theta)sin(\phi)\\
    {y}_p + Lsin(\theta)cos(\phi)\\
    Lcos(\theta)\\
\end{matrix} \right] \quad\text{if,}   0<\theta<\pi \\\\\
 \left[\begin{matrix} 
    {x}_p - Lsin(\theta)sin(\phi + \delta\phi)\\
    {y}_p + Lsin(\theta)cos(\phi + \delta\phi)\\
    Lcos(\theta)\\
\end{matrix} \right] \quad\text{if,}   \pi<\theta<2\pi
\end{cases}
\end{equation} 
\end{footnotesize}

\subsection{3D Reconstruction Along Movement}
The incremental 2D scan along movement can build up a 3D map, the initial position is in the same global frame as that of the 3D scan at static locations. Movement is decomposed into rotation and translation. The rotation is estimated by the two times' fusion, the first time fusion of accelerometer, magnetometer, gyroscope is done at front-end by Madgwick algorithm, the second time fusion of HectorSLAM and Madgwick estimation is completed on laptop through Covariance Intersection(CI), while the position is estimated only by HectorSLAM. Two-level fusion will minimize the uncertainty of the state estimation. The correlation between the two types of yaw angle estimation sources, Madgwick and HectorSLAM is unknown, so CI is applied for the secondary fusion. ($\mu$, $P$) is the estimated mean angle and variance of Madgwick, ($\mu'$, $Q$) is the estimated mean angle and variance of HectorSLAM. The fused result is shown below, $\omega$ is inversely proportional to $P$ and bounded from 0 to 0.5.

\begin{scriptsize}
\begin{equation}\label{eq:ci}
\hat{P}^{-1} = (1-\omega)P^{-1} + \omega Q^{-1},  \omega\in(0,0.5)
\end{equation} 
\begin{equation}\label{eq:ci2}
\hat{\mu} = \hat{P}((1-\omega)\hat{P}^{-1}\mu + \omega\cdot Q^{-1}{\mu'})^{-1}
\end{equation} 
\end{scriptsize}

\begin{figure}[bhtp]
\centering
\includegraphics[scale=.18]{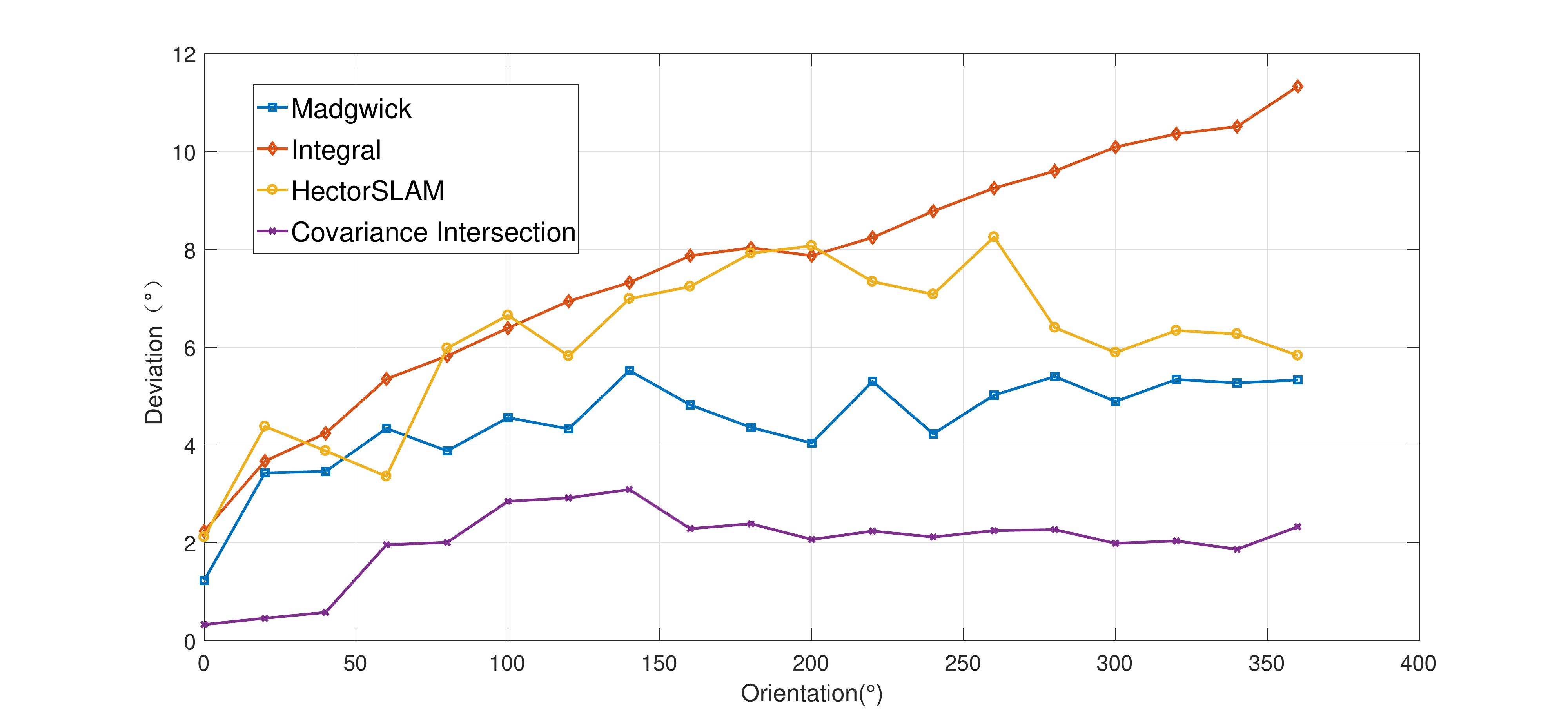}
\caption{Fusion accuracy comparison}
\label{fig:fusion-acc}
\end{figure}

The Fig. \ref{fig:fusion-acc} is the comparison result of yaw angle estimation. “Integral” denotes estimate only depending on gyroscope, a constantly increasing drift over the output presents, which is the worst. The deviation after CI fusion is significantly reduced by half compared to that of a single source.

\begin{figure}[bhtp]
\centering
\includegraphics[scale=.13]{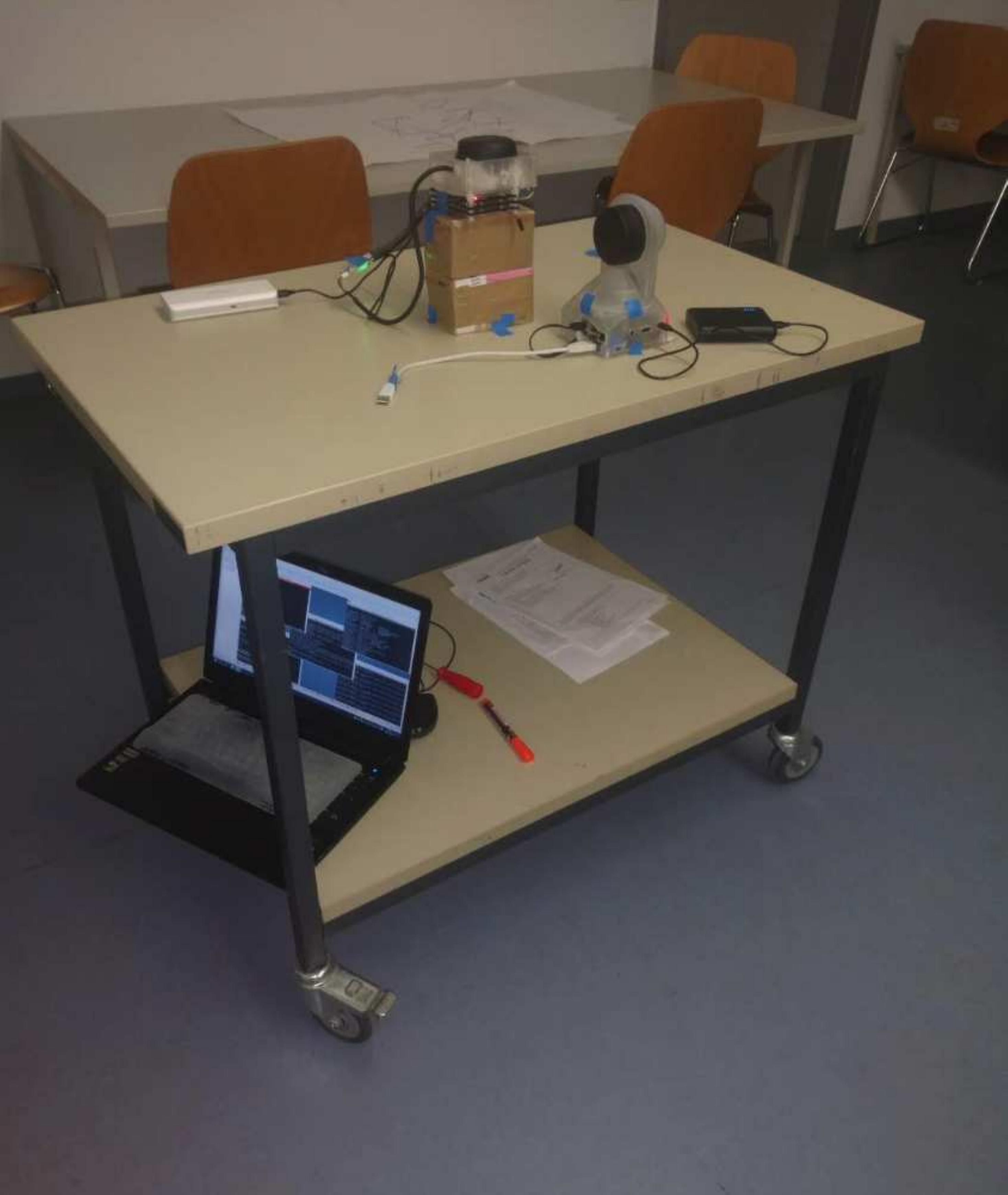}
\caption{Trolley outfitted with sensors}
\label{fig:trolley}
\end{figure}

All synchronization between different sensor topics is based on the time stamp at which the data is sampled. Because both Lidar sensors either for mapping or localization work at low rotation frequency, the entire trolley outfitted with all sensors in Fig. \ref{fig:trolley} can only be pushed along an arbitrary trajectory slowly in the experiment.

\section{METRIC}
In practice, 3D point cloud based evaluation is a difficult task. Here several statistical metrics are proposed, which are all implemented in octree structure. These metrics will evaluate the similarity and difference between two maps quantitatively.\\ 

\subsection{IoU}
The intersection over union is a method, applied in two dimensional visual field like the work in\cite{IoU}, the principle behind it is to find the common part from two compared 3D maps, then divided by union part from two maps. Mark "no" in the following equations indicates the unexplored areas, and the symbol "occ" is short for "occupied" .
The proportions of three types of nodes in the whole octree are calculated. To count unknown nodes, a pre-defined bounding box with known length extracted from octree is employed, then the inner nodes with null pointers in octree are traversed by nested loops. A built-in iterator tool from octomap package is provided for traversing of leaf nodes.

\begin{scriptsize}
\begin{equation}\label{eq:M1}
R_{occ} = \frac{N_{occ}}{N_{no} + N_{occ} + N_{free}}
\end{equation}
\begin{equation}\label{eq:M2}
R_{free} = \frac{N_{free}}{N_{no} + N_{occ} + N_{free}}
\end{equation}
\begin{equation}\label{eq:M3}
R_{no} = \frac{N_{no}}{N_{no} + N_{occ} + N_{free}}
\end{equation}
\end{scriptsize}

Then the three IoU results corresponding to the three types of nodes are derived below. The individual $IoU$ metric corresponding to one of three voxel types is calculated, the "Intersection" indicates the number of voxels with same visualized type in two octomaps, $Union$ corresponds to the total number of the voxels with same type from the compared octomap or the reference octomap. 

\begin{scriptsize}
\begin{equation}\label{eq:M4}
{IoU}_{occ} = \frac{{Intersection}_{occ}}{{Union}_{occ}}
\end{equation}
\begin{equation}\label{eq:M5}
{IoU}_{free} = \frac{{Intersection}_{free}}{{Union}_{free}}
\end{equation}
\begin{equation}\label{eq:M6}
{IoU}_{no} = \frac{{Intersection}_{no}}{{Union}_{no}}
\end{equation}
\end{scriptsize}

Then the final weighted sum IoU is as Equation \ref{eq:M7}.

\begin{scriptsize}
\begin{equation}\label{eq:M7}
{IoU}_{} = 
    \begin{cases}
     R_{occ}\times{IoU}_{occ}+R_{free}\times{IoU}_{free}\quad\text{if,}(R_{occ} + R_{free}) \leq 0.10\\
     R_{occ}\times{IoU}_{occ} + R_{free}\times{IoU}_{free} + R_{no}\times{IoU}_{no} \\ \quad\quad\quad\quad\quad\quad\quad\quad\quad\quad\quad\quad\quad\quad\quad\quad\quad\quad\text{if,} (R_{occ} + R_{free}) > 0.10\\
    \end{cases}
\end{equation}
\end{scriptsize}
The threshold here is set according to the proportion of valid measurements in a bounding box, if the map occupies only a small fraction of the whole volume space, then the final outcome of intersection over Union should be determined only by the occupied and free sets, otherwise the measurement difference in two maps cannot contribute to a remarkable difference in the overall result.

\subsection{Log-odds}
This metric is dependent only on occupied and free nodes with probability values. It is derived from machine learning's loss function for training, but instead of the difference is logged, the ratio of probability values from two maps is taken logarithm, at each overlapping spatial voxel. The log odds output of two identical maps is zero. Logarithm is applied to avoid the quotient being constant, when it near to 0 or near to 1.

\begin{scriptsize}
\begin{equation}\label{eq:M8}
    {Err} =  \sum_{i=1}^{N}
    \begin{cases}
     log(\frac{p^{ref}_i}{p^{tar}_i})\times p^{ref}_i\quad\quad\quad\quad\quad \text{if,} p^{ref}_i \geq 0.9999\\
     log(\frac{1-p^{ref}_i}{1-p^{tar}_i})\times(1-p^{ref}_i)\quad\text{if,} p^{ref}_i \leq 0.0001\\
     log(\frac{1-p^{ref}_i}{1-p^{tar}_i})\times(1-p^{ref}_i)+log(\frac{p^{ref}_i}{p^{tar}_i})*p^{ref}_i\\ \quad\quad\quad\quad\quad\quad\quad\quad\quad\quad\quad\quad\quad\quad\text{if,} 0.0001 < p_{ref} < 0.9999\\         
    \end{cases}
\end{equation}
\end{scriptsize}

The $i$ in Equation \ref{eq:M8} is the index for each node, $N$ is the total number of nodes in the octree. The $ref$ symbol stands for the reference, while $tar$ denotes the target to be compared, The larger this metric value is, the more two octomaps vary.

\subsection{Correlation}
This metric is also borrowed from visual filed, the $normalized$ $cross$ $correlation$ for feature descriptor. This metric is also based on occupied and free nodes only.

\begin{scriptsize}
\begin{equation}\label{eq:M9}
\rho=\\\frac{\sum_{x_0, y_0, z_0}^{x_N, y_N, z_N}\vert (p^{ref}_{x_i, y_i, z_i}-\bar{p})\times(p^{tar}_{x_i, y_i, z_i}-\bar{p})\vert}{\sqrt{\sum_{x_0, y_0, z_0}^{x_N, y_N, z_N} {(p^{ref}_{x_i, y_i, z_i}-\bar{p})}^2\times\sum_{x_0, y_0, z_0}^{x_N, y_N, z_N} {(p^{tar}_{x_i, y_i, z_i}-\bar{p})}^2}}\\      
\end{equation}
\end{scriptsize}
 The coordinates of all nodes in two octrees were already transformed into a same global coordinate during 3D reconstruction process, so the evaluation by this metric can be directly implemented on octomaps. In Equation \ref{eq:M9}, $\bar{p}$ is determined by $\sum_{x_0, y_0, z_0}^{x_N, y_N, z_N} \dfrac{p^{tar}_{x_i, y_i, z_i}+p^{ref}_{x_i, y_i, z_i}}{2N}$, this means probability is averaged over the probabilities from spatially overlapping occupied and free nodes, $x_i,y_i, z_i$, as the coordinate of individual node. The larger this rating is, the more correlated the two compared maps are.

\section{EVALUATION RESULTS}
The 3D reconstruction for a conference room were conducted in two ways, as mentioned before. To verify the proposed metrics. We specially constructed three groups of 3D point clouds at static positions, in which each set of point clouds is formed by splicing several point clouds at known locations. Particularly, the mapping Lidar range was set to 10 meters when the rotation frequency set to 1Hz, at single location, while the hardware settings of mapping Lidar, were kept same during the collection process of other point clouds, all with 5 meters' range when 2 Hz rotation frequency was used. Point clouds in Figure 7 were constructed along motion, point clouds in Figure 8 were built-up at a fixed location, the scanner with the minimal rotating rate and the longest range setting. The point clouds in Figure 9 were collected from four locations, in Figure 10 collected from six locations, point clouds in Figure 7, 9, 10 were constructed under same settings, scanner with rotation rate 2Hz and range at 6m. The point clouds in the leftmost column of Figure 7 to Figure 10 below are original. In the middle column are post-processed point clouds. The rightmost column are the octomaps converted from adjacent filtered point clouds. Here all octrees were built up in the same voxel size. Because the point clouds in these maps are with different characteristics, consequently different PCL filter pipelines were utilized to process them. E.g., point cloud in first row is very sparse, hence there is no need of using down-sampling filter.
\begin{figure}[thbp]%
\centering
\includegraphics[scale = 0.43, angle=90]{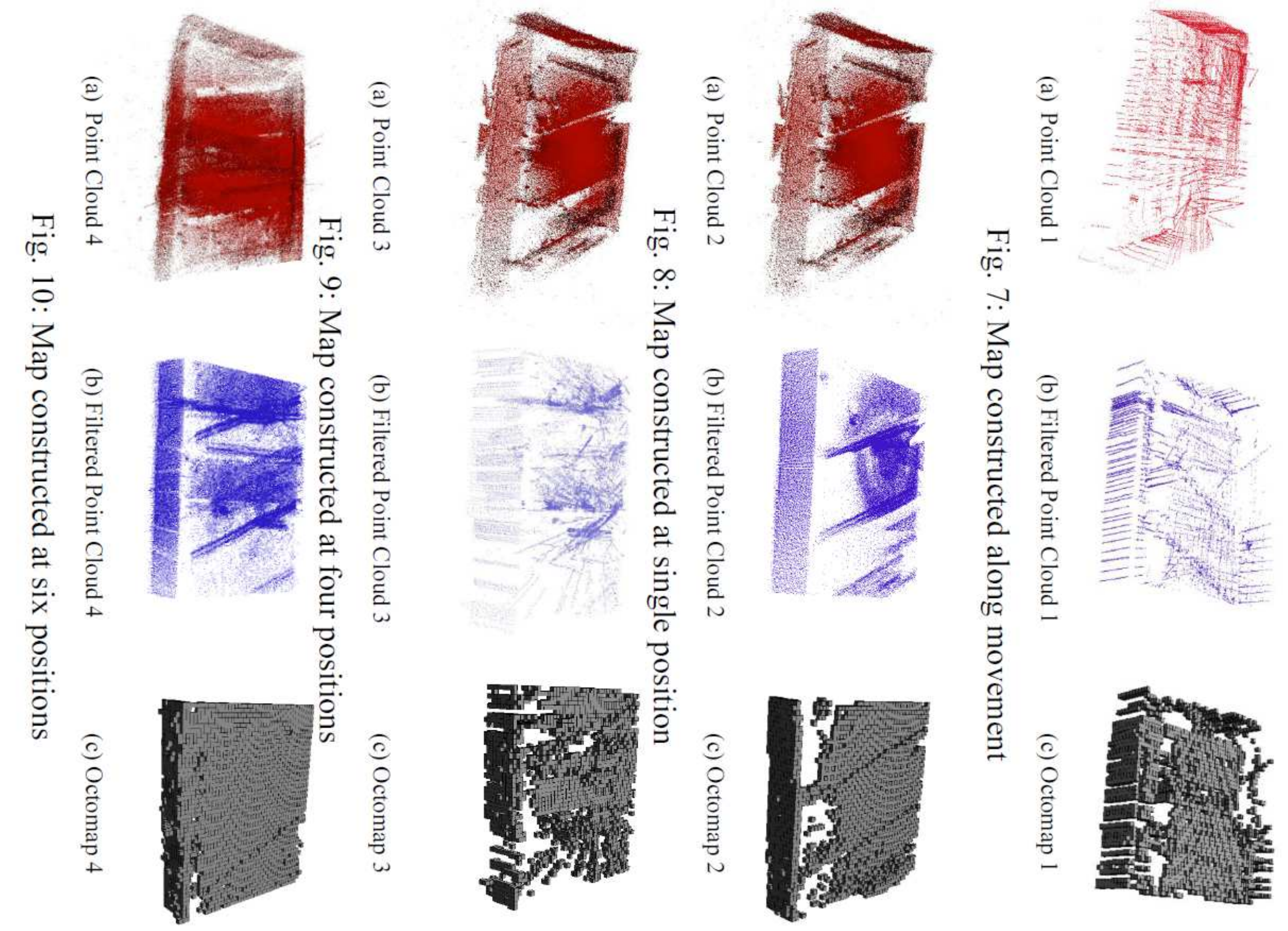}
\label{fig:maps}%
\end{figure}

In the following, symbol "map1" refers to point clouds in Figure 7, others and so on, but the "ref" is used for the point clouds built-up at six locations in Figure 10, which serves as reference for comparison. The Table \ref{tab:Table post-comp} presents adopted PCL filters for individual map. All point clouds went through the down-sampling and pass-through filters, but the point clouds collected from six positions was furthermore filtered by Gaussian filter, because the point clouds collected at six static positions have the most valid points. The pass-through filter intercepts partial point clouds from the original map, to remove the part containing the glass wall. Since the effect of glass wall on measurement is unpredictable. 

\begin{table}[thpb]
  \caption{Post-processing comparison} 
  \label{tab:Table post-comp}
 \begin{center}
  \begin{tabular}{|c|c|c|c|}
  \hline
  &\textbf{Down-sampling}& \textbf{Pass-through} &\textbf{Gaussian}\\
  \hline
  \textbf{Point Cloud 1}& $\times$ & $\checkmark$ & $\times$ \\
  \hline
  \textbf{Point Cloud 2}& $\checkmark$ & $\checkmark$ & $\times$ \\
  \hline
  \textbf{Point Cloud 3}& $\checkmark$ & $\checkmark$ & $\times$ \\
  \hline
  \textbf{Point Cloud 4}& $\checkmark$ & $\checkmark$ & $\checkmark$ \\
  \hline
  \end{tabular}
 \end{center}
\end{table}

Table \ref{tab:Table 6} makes a basic statistical analysis of maps, unknown nodes are not taken into consideration in our case, because the point cloud filtered by pass-through filter only retains most of the valid measurements in the known area. \\
\begin{table}[thpb]
  \caption{Proportion} 
  \label{tab:Table 6}
 \begin{center}
  \begin{tabular}{|c|c|c|c|}
  \hline
  &\textbf{Occupied Ratio}& \textbf{Free Ratio} &\textbf{Leaf nodes number}\\
  \hline
  \textbf{map1}& 13.059\% & 86.941\% & 17140 \\
  \hline
  \textbf{map2}& 19.061\% & 80.939\% & 18320\\
  \hline
  \textbf{map3}& 18.6405\% & 81.3595\% & 20553\\  
  \hline
  \textbf{ref}& 30.4077\% & 69.5923\% & 22491\\  
  \hline
  \end{tabular}
  \end{center}
\end{table}

Proportions of three compared octomaps are almost same, but they are different from that of reference map, so it is very hard to rate the compared octomaps by this rough method. Particularly, the point clouds reconstructed incrementally along movement are very sparse, consequently the generated octomap is with many defects. The figure below shows the final scores via our metrics, they are all normalized ranging in 0 to 1, map1 to map3 are all compared against reference map. The ideal result is at rightmost for the two same maps, log odds is 0, IoU metric and Correlation metric score 1. Here the log odds is an average value over all the free and occupied nodes. The additional histogram in red is the mean probability of common nodes in the two compared octomaps, and the vertical bar on top is the average probability deviation. The final result shows that, the octomap generated from point cloud at a single location is the most consistent, followed by the octomap from point clouds at four locations, the octomap from point clouds collected along movement is the worst. 

   \begin{figure}[thpb]
     \centering
      \includegraphics[scale = 0.40]{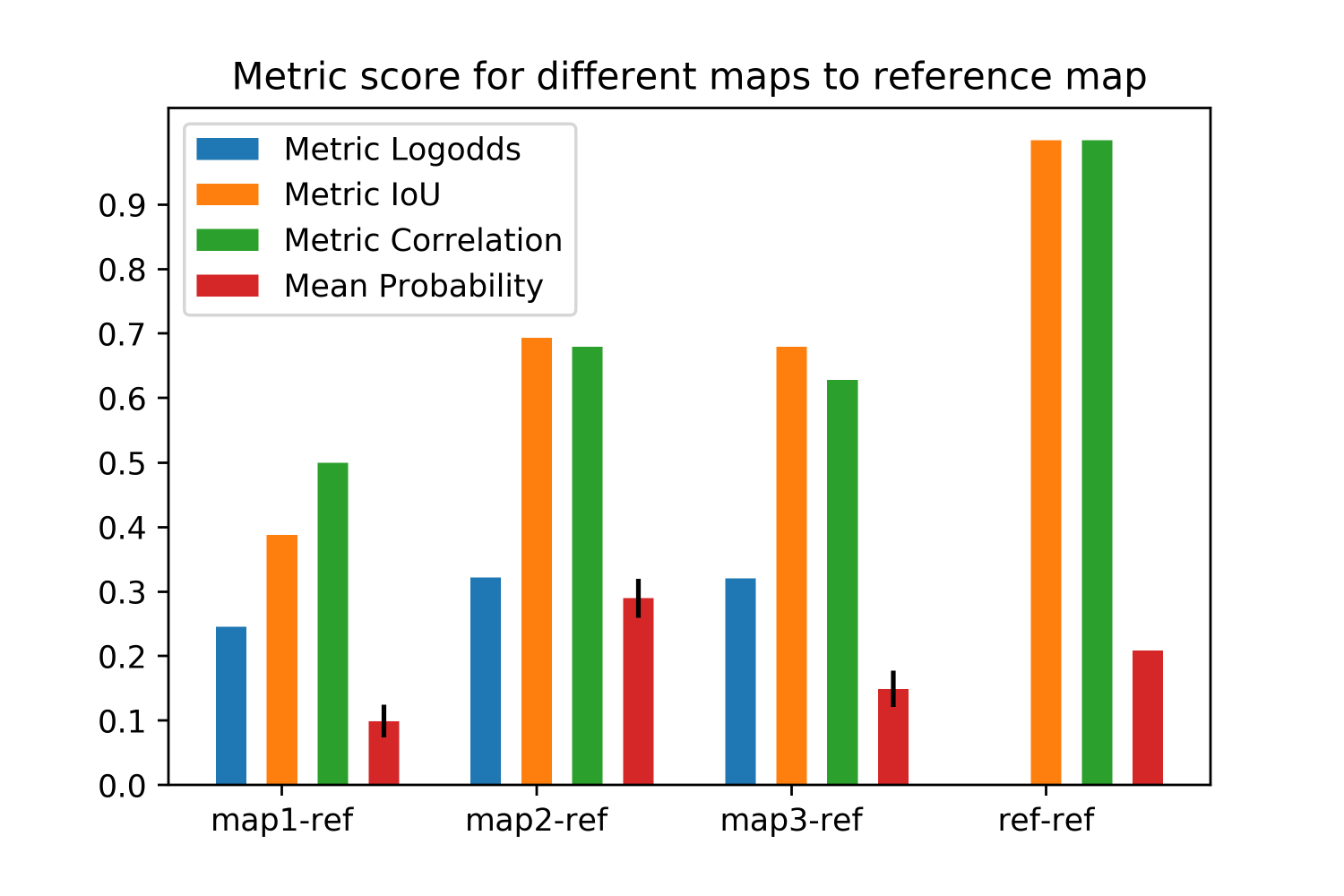}    
      \caption*{Fig. 11: Metric score}
      \label{fig:metrics}
   \end{figure}

The leaf node with the size attribute in octree is visualized in octomap, and its size is determined by resolution setting. The following figure displays the required time to convert the point cloud to the octomap, and the occupied volume at different resolutions of octree. The whole evaluation process, including all three metrics above is implemented on two exactly same octomaps at different resolutions. The smaller the size of voxel in the octree is, the smoother the reconstructed structure is. The big size of the occupied cells will make the map with many enclosures, unable to be used for navigation. However, the higher resolution increases the whole evaluation time, so a trade-off between computational time and map quality should be found. The octomap at resolution 0.20m, built-up at six positions is with 9979 nodes in the octree, the evaluation time takes only 75ms, therefore the metric can be implemented in real time potentially.

\begin{figure}[thbp]%
\centering
\includegraphics[scale = 0.38]{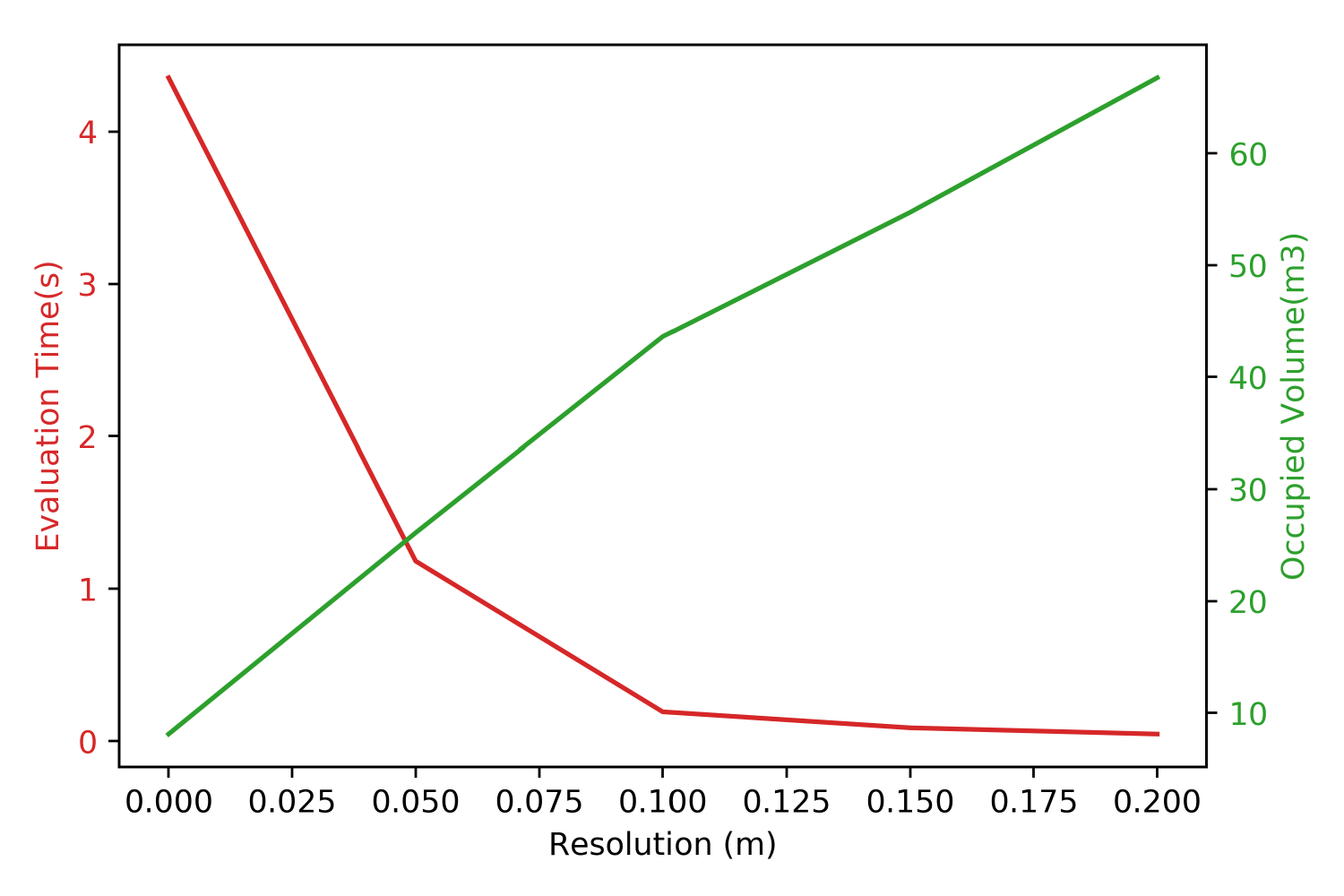}
\caption*{Fig. 12: Relationship between resolution and size, time}
\label{fig:resolution}
\end{figure}

\section{CONCLUSIONS}
Through the experimental tests, the entire system can achieve the 3D reconstruction by two means, including incremental 2D scan along movement, or the 3D scan completed at fixed locations. The octomap based metrics can assess the target and reference maps' similarity and difference comprehensively. The final test, also indicates that the evaluation by the proposed metrics, can be completed in hundred milliseconds level, but constrained by other parameters, like the total number of points and the voxel size. With appropriate parameters, the metrics can be implemented while constructing the octomap incrementally, because 2D Lidar's maximal sampling rate is around 1000 points per second, the number of measurements in this order of magnitude, can be converted to octree nodes within a few milliseconds by ray-casting, so the total time including conversion and assessment process can be performed within 1 second. At the same time, we should be aware that, the use of low cost 2D Lidar sensors off-the-shelf to build 3D point cloud, will either increase the complexity of the hardware, like the scan with additional dimension controlled by motor, or requires additional ego-motion estimation sensor, which increases the complexity of the software. Because the rotation frequency of the low-cost Lidar is not high, therefore, the whole frame work is not applicable to the mapping in a fast and continuous motion. The complete 3D scan process relying on 2D Lidar is also quite time-consuming, making the whole system for 3D reconstruction only applicable to some low-speed mobile platforms to preform 3D perception.
\addtolength{\textheight}{-12cm}  







\end{document}